\pdfoutput=1

\documentclass[11pt]{article}

\usepackage[]{acl}
\usepackage[inkscapelatex=false]{svg}

\usepackage{times}
\usepackage{latexsym}
\usepackage{amssymb}
\usepackage{multirow}
\usepackage{multicol}
\usepackage{booktabs}
\usepackage{soul}
\usepackage{xcolor}
\usepackage{colortbl}
\usepackage{listings}
\usepackage{courier}
\usepackage{url}
\usepackage{sourcecodepro}
\usepackage[most]{tcolorbox}
\usepackage{listings}
\usepackage{fancybox}
\usepackage{enumitem}
\usepackage{lipsum}

\usepackage{adjustbox}

\usepackage[T1]{fontenc}

\usepackage[utf8]{inputenc}

\usepackage{microtype}

\usepackage{inconsolata}

\usepackage{pgfplots}
\usepgfplotslibrary{groupplots}
\usetikzlibrary{patterns}

\usepackage{ulem}

\definecolor{C_BASE}{HTML}{007694}
\definecolor{C_BASE_SOFT}{HTML}{66adbf}
\definecolor{C_BASE_SOFTER}{HTML}{b9e5f0}

\definecolor{C_SECOND}{HTML}{B62699}
\definecolor{C_SECOND_SOFT}{HTML}{e2a8d6}
\definecolor{C_SECOND_SOFTER}{HTML}{f2c7ea}

\definecolor{C_THIRD}{HTML}{00B9E8}
\definecolor{C_FOURTH}{HTML}{0B008F}
\definecolor{C_FOURTH_SOFTER}{HTML}{b9b6e3}
\definecolor{C_BACKGROUND}{HTML}{EDF5FC}

\usepackage{pgfplotstable}
\pgfplotsset{every tick label/.append style={font=\footnotesize},compat=1.17}
\pgfplotsset{compat = newest}

\title{BizBench: A Quantitative Reasoning Benchmark for Business and Finance}

\author{Rik Koncel-Kedziorski\textsuperscript{\textdagger}, Michael Krumdick\textsuperscript{\textdagger}\\\textbf{Viet Lai, Varshini Reddy, Charles Lovering, Chris Tanner} \\
  Kensho Technologies\\
  \texttt{\{rikka, michael.krumdick\}@kensho.com}
  }

\begin{document}
\maketitle
\begin{abstract}

Answering questions within business and finance requires reasoning, precision, and a wide-breadth of technical knowledge. Together, these requirements make this domain difficult for large language models (LLMs). We introduce BizBench, a benchmark for evaluating models' ability to reason about realistic financial problems. BizBench comprises eight quantitative reasoning tasks, focusing on question-answering (QA) over financial data via program synthesis. We include three financially-themed code-generation tasks from newly collected and augmented QA data. Additionally, we isolate the reasoning capabilities required for financial QA: reading comprehension of financial text and tables for extracting intermediate values, and understanding financial concepts and formulas needed to calculate complex solutions. Collectively, these tasks evaluate a model's financial background knowledge, ability to parse financial documents, and capacity to solve problems with code. We conduct an in-depth evaluation of open-source and commercial LLMs, comparing and contrasting the behavior of code-focused and language-focused models. We demonstrate that the current bottleneck in performance is due to LLMs' limited business and financial understanding, highlighting the value of a challenging benchmark for quantitative reasoning within this domain.


\end{abstract}

\newcommand{\mtr}[2]{\multirow{#1}{*}{\textbf{#2}}}
\newcommand{\mtc}[2]{\multicolumn{#1}{c}{\textbf{#2}}}
\newcommand{\mtcb}[2]{\multicolumn{#1}{|c|}{\textbf{#2}}} 
\newcommand{\mrt}[2]{\multirow{#1}{*}{\rotatebox{90}{#2}}}

\newcommand{\hlr}[2]{\colorbox{red}{$\displaystyle #1$}}
\newcommand{\hly}[2]{\colorbox{yellow}{$\displaystyle #1$}}
\newcommand{\hlc}[2]{\colorbox{cyan}{$\displaystyle #1$}}

\definecolor{ks1}{HTML}{007694}
\definecolor{ks2}{HTML}{B62699}
\definecolor{ks3}{HTML}{00B9E8}
\definecolor{ks4}{HTML}{0B008F}
\definecolor{commentgreen}{HTML}{459a4b}

\lstset{language=Python,
    backgroundcolor=\color{ks4!10},
    basicstyle=\scriptsize\ttfamily,
    rulecolor=\color{black},
    commentstyle=\color{commentgreen}\ttfamily,
    keywordstyle=\color{blue}\bfseries, 
    upquote=false,
    numbers=left,
    numberstyle=\tiny\color{gray},
    stepnumber=1,
    numbersep=-3pt,
    showstringspaces=false,
    breaklines=true,
    frameround=ftff,
    frame=none,
    belowcaptionskip=0em,
    belowskip=0em
}

\section{Introduction}

Large language models (LLMs) show strong performance on question-answering (QA) and code generation tasks \citep{austin2021program, openai2023gpt4}. Nonetheless, it remains particularly difficult for models to reason about quantities and numbers \citep{hendrycksmath2021}. This poses issues for using LLMs for real-world problems in business and finance, as these fields can require transparent and precise reasoning capabilities.

\begin{figure}[t]
    \centering
    \includegraphics[width=0.9\linewidth]{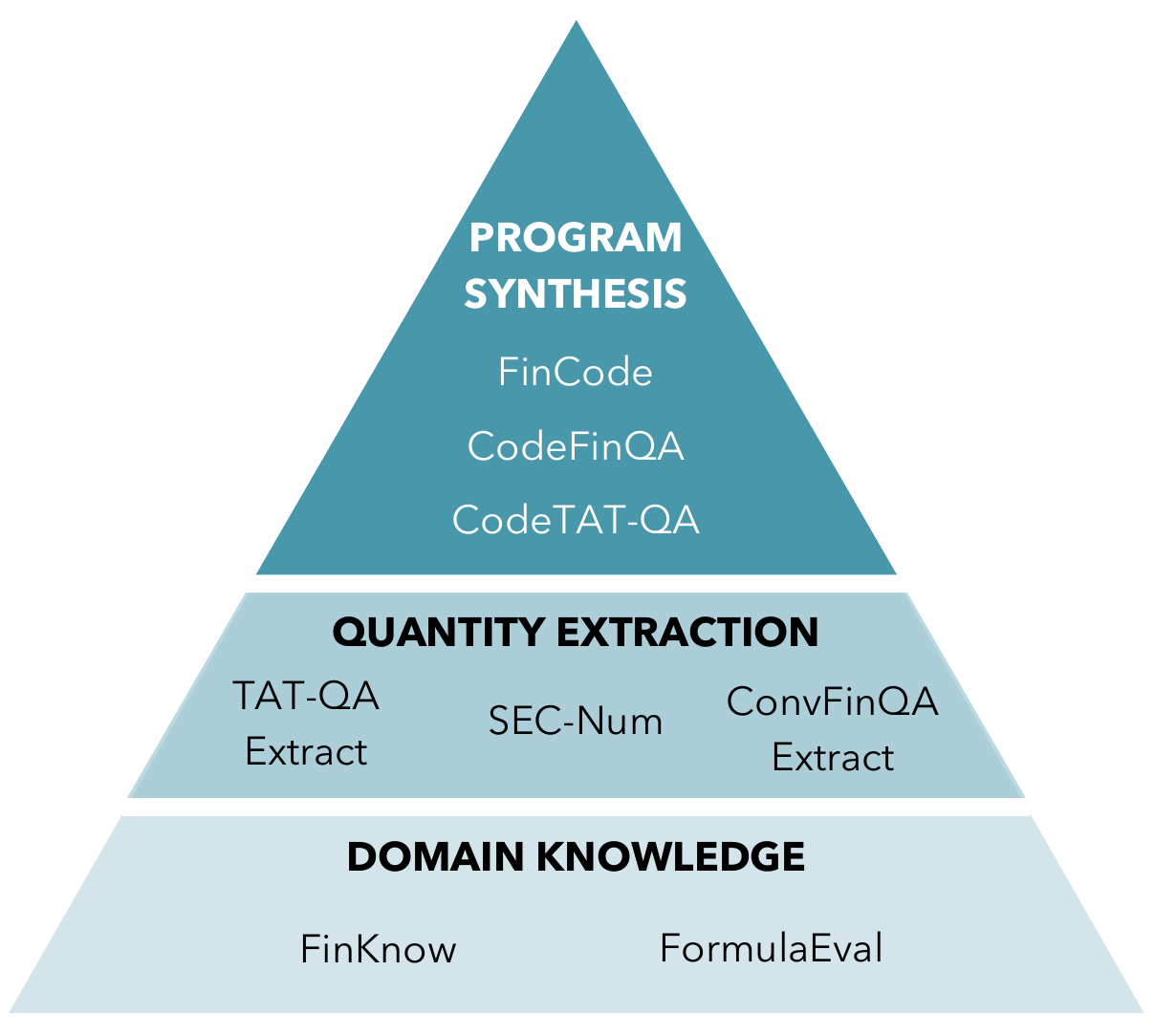}
    \caption{Overview of BizBench’s eight tasks. Each level of the pyramid corresponds to a task category, with higher levels requiring increasingly complex capabilities.}
    \label{fig:pyramid}
\end{figure}

To facilitate the development of better models for business and finance, we introduce a new benchmark for evaluating financial quantitative reasoning, with a focus on question-answering over structured and unstructured financial data. 

Financial questions often require multi-step reasoning \citep{chen-etal-2021-finqa}. 
Professional workflows necessitate a transparent reasoning process to promote user trust, but how LLMs reason remains opaque. 
While chain-of-thought (CoT) prompting, where reasoning steps are generated as part of model output, improves performance on reasoning tasks \citep{suzgun-etal-2023-challenging,kojima2022large}, the answers generated from CoT are often not a direct product of the generated reasoning \citep{Wei2022ChainOT,wang-etal-2023-towards,lanham2023measuring}.
Unlike generating CoT, generating executable code (program synthesis) ties model outputs to
specific opereations, functions or instructions \citep{madaan-etal-2022-language,chen2022program}. In BizBench, we frame multi-step QA as a program synthesis task. This formulation provides increased transparency because the exact rationale for a model's answer can be audited.

BizBench consists of three interrelated \textit{types} of tasks for assessing transparent and accurate financial reasoning: program synthesis, quantity extraction, and domain knowledge. Figure~\ref{fig:pyramid} shows an overview of BizBench.

{\it Program Synthesis.} We introduce a novel QA task, \textbf{FinCode}, built from professional exams. We also create two code-generation tasks \textbf{CodeTAT-QA} and \textbf{CodeFinQA} by reformulating the existing financial QA datasets TAT-QA \cite{zhu-etal-2021-tat} and FinQA \cite{chen-etal-2021-finqa}. Both of these tasks require leveraging raw text and tables.

{\it Quantity Extraction.} We introduce \textbf{SEC-Num}, a numerical span-identification task over corporate earnings reports filed with the U.S. Securities and Exchange Commission (SEC). Additionally, we study span-based QA subsets of ConvFinQA \cite{chen-etal-2022-convfinqa} and TAT-QA \cite{zhu-etal-2021-tat}.

{\it Financial Domain Knowledge.} We introduce two tasks: \textbf{FormulaEval}, a novel code-completion task which tests a model's knowledge of financial formulas, and \textbf{FinKnow}, which tests non-quantitative subsets of business and finance exams.

\begin{table}[t]
\label{tab:taskmatrix}
    \centering
    \resizebox{\linewidth}{!}{
    \begin{tabular}{lccccccc}
    \hline
        \mtr{2}{Task} & \mtr{2}{Test} &  \mtr{2}{Train} & {\bf Novel } & \mtr{2}{CG} & \mtr{2}{FK} & \mtr{2}{DC} &\mtr{2}{DS} \\
        &   & & {\bf Task} &  & & &  \\

        \hline
         \multicolumn{6}{l}{\it Program Synthesis} \\
         $\>$ FinCode & 121 & 16 &  $\checkmark$ & $\checkmark$ & $\checkmark$ &  & \\ 
         $\>$ CodeFinQA & 844 & 4,669 & & $\checkmark$ & $\checkmark$ & $\checkmark$ & \\
         $\>$ CodeTAT-QA & 392 & 2,864 &   & $\checkmark$ & $\checkmark$ &  & $\checkmark$\\
         \hline
         \multicolumn{6}{l}{\it Quantity Extraction}  \\

         $\>$ ConvFinQA (E) &  916 & - &  &  & $\checkmark$ & $\checkmark$ & \\ 
         $\>$ TAT-QA (E) & 248 & - & &  & $\checkmark$ & $\checkmark$ &  \\
         $\>$ SEC-Num&  2,000 & 6,845 & $\checkmark$ &  & $\checkmark$ &$\checkmark$  & \\
         \hline
         \multicolumn{6}{l}{\it Domain Knowledge} \\
         $\>$ FinKnow   & 877 & - & & & $\checkmark$  & & \\
         $\>$ FormulaEval & 50 & - & $\checkmark$ & \checkmark & $\checkmark$  & & \\
         \hline

    \end{tabular}
    }
\caption{Characteristics of BizBench tasks including test data size, novel supervised-finetuning data size, the inclusion of novel task and novel data, and if the task involves code generation (CG), financial knowledge (FK), document context (DC), and data structures (DS).}
    \label{tab:task-characteristic}
\end{table}

BizBench complements existing benchmarks for business and financial NLP \cite{xie2023pixiu,shah-etal-2022-flue}. Existing benchmark tasks include sentiment analysis, named entity prediction, and stock price prediction. BizBench focuses on a model's capacity to leverage numeric information in structured and unstructured financial data to answer quantitative financial questions.

Our evaluations of open-source and commercial pre-trained language models -- in both few-shot and fine-tuned setups -- provide a detailed view of the state-of-the-art NLP for quantitative financial reasoning. Model size, instruction-tuning, and code-specific pretraining all meaningfully impact performance, but significant improvement is still needed for even the best models to be useful in high-stakes, real-world workflows.

The major contributions of this work include:
\begin{itemize}[itemsep=2pt,topsep=4pt,parsep=2pt,partopsep=2pt]
    \item A new benchmark of eight tasks for evaluating quantitative reasoning in finance and business.
    \item Novel data for studying the problems of program synthesis, quantity extraction, and knowledge of the financial domain.
    \item Detailed evaluation of a variety of LLMs, showing how model size, data, code-tuning, and alignment impact task performance.
    \item Error analysis showing that top-performing models fall short mainly due to gaps in their financial knowledge.
\end{itemize}
Along with releasing the train, development and unlabeled test sets for BizBench, we've established a leaderboard at \url{anonymized}, open to public submission. Through the leaderboard, we hope to encourage progress towards developing solutions to these challenging problems while mitigating the potential for data contamination.

\section{Reasoning in Business and Finance}
\label{sec:bg}

Business and finance professionals are frequently required to reason about quantities.  
They often search for specific quantities within large sets of distractors (e.g., searching for the revenue of a particular division for a specific quarter, and doing so from a report that contains a comprehensive detailing of financial metrics across many divisions).
They also rely on financial domain knowledge across banking, accounting, and finance to manipulate these quantities in useful ways.
For example, computing an EBITDA margin requires first identifying a series of relevant quantities (net income, interest expense, taxes, depreciation, amortization, and revenue) and then applying a specific formula to these quantities (taking the sum of the first five quantities and computing their percentage of revenue).
Since large sums of money are often at stake, financial professionals need to communicate transparent rationales for their decisions. 

Researchers have developed tasks to measure some aspects of quantitative financial reasoning capabilities of AI systems.
FinQA consists of 8,281 questions written by financial professionals, paired with context and tables from earnings reports of the largest companies listed on American stock exchanges \cite{chen-etal-2021-finqa}. 
Answering FinQA questions often requires multiple reasoning steps, including extracting quantities, performing mathematical operations with them, and reasoning about time periods such as fiscal quarters and years. 
Additionally, these questions test the financial knowledge of currency and scale (e.g. percentage, millions, billions).
Similar to FinQA, TAT-QA consists of questions, answers, and relevant context of text and tables \cite{zhu-etal-2021-tat}.
The 16,552 questions in TAT-QA cover a range of skills including span-extraction, comparison, and arithmetic. 

Program synthesis improves transparency of model outputs, allowing for auditing of reasoning steps --- which in turn increases trust and usability. Since large sums of money are often at stake, financial professionals need to communicate transparent rationales for their decisions. 
 Generating programs allows QA models to avoid arithmetic calculations, which is challenging for smaller models.



\section{Task Details}
\label{sec:tasks}
The main focus of BizBench is evaluating a model's financial understanding through program synthesis. 
We additionally provide tasks in two domains we view to be the building blocks in this regard: quantity extraction and financial domain knowledge.
In order to answer a complex program synthesis question, the model must be able to understand and extract relevant values within a question or context. It  must also have some knowledge of the necessary formulas and other information required to then compute the answer.
Additional skills and features of each task are shown in Table~\ref{tab:task-characteristic}.

\subsection{Program Synthesis}
\label{sec:ps}
Program synthesis requires a model to generate logically sound code that can be executed to answer some question. Each example contains a natural language question, optionally text or structured data source, and a Python program that produces a numeric answer to the question. We measure accuracy by comparing this numeric output with a ground truth reference value. Answers within 1\% of the reference are considered to be correct. 


\newcommand{\listingsttfamily}{\fontfamily{SourceCodePro-TLF}\small}
\lstset{numberstyle=\ttfamily}

\begin{figure}[t]
\small
{\bf Question:} {\it An investment project costing \$500 today will generate profiles of \$500 in year five and year ten. If the discount rate is 10\%, what is the project's NPV? Answer to the nearest dollar.}
\begin{tcolorbox}[enhanced jigsaw, colback=white, left=0pt, right=0pt, top=0pt, bottom=0pt, title=Python Program Solution]
\vspace{-2mm}
\begin{lstlisting}[backgroundcolor=\color{white},
    rulecolor=\color{black},
    rulesepcolor=\color{gray},
    % numbers=none,
    ]
  initial_investment = 500
  discount_rate = 0.1
  pv = lambda year, cash_flow:                 cash_flow / ((1 + discount_rate) ** year)
  npv = (pv(5, 500) + pv(10, 500)) - initial_investment
  round(npv)
\end{lstlisting} 
\end{tcolorbox}
\caption{Example from {\bf FinCode}. In this dataset, answering questions requires financial background knowledge and the ability to synthesize code.}
\label{fig:fincode-example}
\end{figure}

Questions for these tasks are written by financial professionals. The associated reference code is written either by human annotators or converted from human-generated equations (see Figure~\ref{fig:codefinqa}) for a sample semantically-rich Python program. 

\begin{table}[ht]
\centering
\resizebox{\linewidth}{!}{
    \begin{tabular}{lrrr}
        \hline
        \mtr{1}{Indicator} & \mtr{1}{FinCode} & \mtc{1}{CodeFinQA} & \mtc{1}{CodeTAT-QA} \\ 
         \hline
         Context  & - & 131.3 & - \\
         Table & - & 64.8& 147.0\\
         Question & 14.4 & 5.4 & 6.7 \\
        \textit{Avg. Total Numerals} & 14.4 & 201.6 & 153.7 \\
     
        \hline
        Addition          & 0.81  & 0.25  & 0.13 \\
        Subtraction       & 1.04  & 0.49  & 2.43 \\
        Multiplication    & 1.40  & 0.62  & 0.41 \\
        Division          & 0.73  & 0.73  & 0.43 \\
        Percentage        & 0.21  & 0.62  & 0.39 \\
        Exponent          & 0.13  & 0.00  & 0.00 \\
        \textit{Avg. Total Operators} & 4.32 & 2.71 &  3.79 \\
        \hline
        Lines of code    & 7.0 & 3.8 & 3.0 \\
         Parenthesis     & 4.6  & 0.3  & 1.6 \\
         \hline
    \end{tabular}
}
\caption{Complexity of program synthesis in BizBench including the average count of numerals within each part of the input;  the average count of mathematical operators used in the solution; and the solution complexity in terms of average lines of code and parentheses;.}
\label{tab:code-complexity}
\end{table}

We highlight some important features of these tasks: First, the questions are written by financial professionals using real-world data and financial knowledge. As such, they are closer to the kinds of questions that business and financial professionals answer as part of their workflows. They present different challenges from questions found in existing numerical reasoning datasets because they involve resolving complex quantity references, avoiding abundant distractor quantities, and using implicit financial background knowledge. These tasks are different from existing code generation tasks in that they require grounding generated code in real-world text or data structure context that also require financial background knowledge. 

Secondly, the code we provide with these tasks is {\it semantically-rich}, by which we mean that the relationship between the code, the question it answers, and the context it uses is intuitive. 
This is accomplished through the use of a simple procedural style and descriptive variable names. 
Semantically-rich code explains the model's final answer, allowing model answers to be audited. Although we do not enforce the generation of semantically rich code, we find that in practice providing the model with semantically rich examples is sufficient to elicit this behavior.

Further details of the data collection process for the following tasks can be found in Appendix~\ref{app:data-collection}.

\textbf{FinCode} consists of 137 questions, programs, and answers taken from Certified Financial Analyst (CFA) and Certified Public Accountant (CPA) practice exams.
Each question is annotated with Python code that references quantities from the question text and applies mathematical operations to compute the requested numeric answer. 
The financial background knowledge required to answer these questions is requisite for professionals earning certification. In total, 46 of the 137 examples were written from scratch by financial professionals, and the remaining 91 were generated by an LLM and then verified by financial professionals.


Figure~\ref{fig:fincode-example} shows a typical example from this dataset. 
The question statement discusses a potential investment project. 
Answering this question requires understanding domain-specific terms -- such as ``discount rate'' and ``NPV'' (net present value) -- and how the concepts fit together into a formula for determining the answer. 
The code to answer this question is complex, requiring multiple steps of financial arithmetic. 
See Table~\ref{tab:code-complexity} for an analysis of the complexity of FinCode and the other BizBench datasets. 


\begin{figure}[t]
\small
{\bf Context:} {\it ``... Q4 2022 revenue totaled 28.9B, compare to the same period last year of 27.8B. We saw stronger sales in our leasing division with a 14\% increase ...''}\\
{\bf Question:} {\it What was the percent change in revenue from 2021 to 2022?''}
\begin{tcolorbox}[enhanced jigsaw, colback=white, left=0pt, right=0pt, top=0pt, bottom=0pt, title=Original FinQA derivation]
\vspace{-2mm}
\begin{lstlisting}[backgroundcolor=\color{white},
    rulecolor=\color{black},
    rulesepcolor=\color{gray},
    ]
  divide(subtract(28.9, 27.8), 27.8)
\end{lstlisting}
\end{tcolorbox}
\begin{tcolorbox}[enhanced jigsaw, colback=white, left=0pt, right=0pt, top=0pt, bottom=0pt, title=Our Python Program]
\vspace{-2mm}
\begin{lstlisting}[backgroundcolor=\color{white},
    rulecolor=\color{black},
    rulesepcolor=\color{gray},
    % numbers=none,
    ]
  revenue_2022 = 28.9
  revenue_2021 = 27.8
  change = revenue_2022 - revenue_2021
  percent_change = change / revenue_2021
  answer = percent_change * 100
\end{lstlisting} 
\end{tcolorbox}
\caption{{\bf CodeFinQA} example. Comparison of our added code solution and the original FinQA equation. Our Python program is executable with named variables for easier verification (lines 1-2),  quantity composition (lines 3-4), and with the expected scale or unit (line 5).}

\label{fig:codefinqa}
\end{figure}

\textbf{CodeFinQA} is a subset of FinQA for which we provide code solutions to the questions. 
This dataset of 5,513 question/context/code triples can be used for finetuning or dynamic prompting.
For the CodeFinQA task, models are given the same input as FinQA: text, table, and the question. 
The output is Python code which is executed to produce an answer. A sample is shown in Figure~\ref{fig:codefinqa}.

\textbf{CodeTAT-QA} is a subset of QA pairs from the TAT-QA dataset 
that can be answered using information in the provided table.
We adapt this dataset for QA over structured financial data, a common task in business and finance workflows. 
We believe QA models can benefit from learning how to access these data sources programmatically. 

Each question is augmented not only with a program, but also with a structured table representation that the model can utilize to compute its solution.
Information from the table can be accessed through the dataframe by specifying the row and column labels, as shown in Figure~\ref{fig:codetatqa-example}.





\lstset{numberstyle=\ttfamily}

\begin{figure}[t]
\small
{\textbf{Question:} \textit{What was the change in Foreign revenue in 2019 from 2018?}}

\centering
\begin{tabular}{|l|r|r|r|}
    \mtc{4}{Balance Sheet} \\
    \hline
    \mtcb{1}{Revenue, in millions} & \mtcb{1}{2019} & \mtcb{1}{2018}      & \mtcb{1}{2017} \\
    \hline
    Domestic  & 204.2    & 140.3   & 56.0 \\
    Foreign       & \colorbox{C_SECOND_SOFTER}{11.8}   & \colorbox{C_BASE_SOFTER}{19.9}   & 14.2  \\
    Income before income taxes     & {216.0}      & 160.2      & 70.2      \\
    \hline
\end{tabular}

\begin{tcolorbox}[enhanced jigsaw, colback=white, left=0pt, right=0pt, top=0pt, bottom=0pt, title=Dataframe Access]
\vspace{-2mm}
\begin{lstlisting}[backgroundcolor=\color{white},
    rulecolor=\color{black},
    rulesepcolor=\color{gray},
    % numbers=none,
    escapechar=!
    ]
  df = DataFrame(data=balance_sheet_table)
  foreign_2019 = !\colorbox{C_SECOND_SOFTER}{df["Foreign"]["2019"]}!
  foreign_2018 = !\colorbox{C_BASE_SOFTER}{df["Foreign"]["2018"]}!
  answer = foreign_2019 - foreign_2018
\end{lstlisting}
\end{tcolorbox}
\caption{Example from \textbf{CodeTAT-QA}. The task requires accessing and manipulating data via a dataframe, simulating QA scenarios where data comes from structured sources.}
\label{fig:codetatqa-example}
\end{figure}

\subsection{Quantity Extraction}
Quantity extraction tasks require models to identify numbers in text and tables from natural language descriptions. 
Although this task can sometimes require minimal financial background knowledge (e.g. reading a number directly from a table), it is a necessary sub-task for many complex tasks that do. 
It is also a valuable task in its own right, as many business and finance workflows can benefit from high precision automated quantity extraction.  
BizBench includes three quantity extraction tasks: a new dataset of SEC filings and labeled quantities (SEC-Num), and extraction-only subsets of TAT-QA and ConvFinQA.

\textbf{SEC-Num} is a novel dataset for quantity extraction from SEC filings. 
Recently, the SEC implemented a machine-readable labeling scheme for structuring data within human-readable documents.\footnote{\url{https://www.sec.gov/structureddata/osd-inline-xbrl.html}}
Under these rules, filers are required to annotate quantities within reports with natural language descriptions of each quantity reported. 
We treat these descriptions as labels and define the SEC-Num task as follows: given a document snippet and a target label as input, the expected output is the quantity span from the snippet corresponding to the label.
This open-vocabulary task generalizes \citet{loukas-etal-2022-finer}, who focus on the most frequent labels and develop a classification task. 
A snippet of the original SEC filing is shown in Table~\ref{tab:appendix-sec-filing} in the Appendix.

The data processing pipeline for SEC-Num begins with 202 10-K and 10-Q filings from the SEC EDGAR data portal. 
From these, we split each document into pages, each of which may contain multiple paragraphs and tables with a large number of quantities.
For each unambiguous quantity label, we create a datapoint $(x,y)$ where $x$ is a snippet/label pair and $y$ is the corresponding number from the snippet for the given label. 
The resulting dataset has 8,845 datapoints, which we split into 6,845 train and 2,000 test datapoints. 
Full statistics of this data are available in Table~\ref{tab:task-characteristic}. 

\textbf{TAT-QA Extract (E)} and \textbf{ConvFinQA Extract (E)} are subsets of questions from TAT-QA  and ConvFinQA respectively, which can be answered using a numeric span from the context text or tables.
See Table~\ref{tab:task-characteristic} for our dataset statistics.

\subsection{Domain Knowledge}
These tasks test the financial domain knowledge of an AI system.
Here, models must demonstrate internal understanding of business and financial terms, practices, and formulae. 

\textbf{FinKnow} contains 877 multiple choice questions and answers collected from CFA practice exams and the business ethics, microeconomics, and professional accounting exams from the MMLU dataset \cite{hendryckstest2021}. The CFA exam questions have three potential choices, while the questions from the MMLU dataset have four.
We exclude incomplete questions and questions that require numeric extraction or numerical reasoning.
In total, this dataset contains 418 CFA, 86 business ethics, 224 microeconomics and 149 professional accounting question-answer pairs. We evaluate the models in a zero-shot setup. For each question, we compute the log probability of each potential answer and select the highest as the model's choice. 

\textbf{FormulaEval} is a novel code-completion task designed to determine whether formulae for different business, economic, and financial measures are memorized and accessible without external knowledge sources.
Using these formulae is required for the program synthesis tasks and is an important part of many business and financial workflow.
\begin{table*}[t!]
\resizebox{\textwidth}{!}{
\begin{tabular}{lrccccccccc}
\hline
    \mtr{3.5}{Model} & \mtr{3.5}{Size} & \mtc{2}{Domain Knowledge}  & \mtc{3}{Quantity Extraction} & \mtc{3}{Program Synthesis} & \mtr{3}{Avg.}\\
    
    \cmidrule(lr){3-4}
    \cmidrule(lr){5-7}
    \cmidrule(lr){8-10}
    &  & {\bf FinKnow} & {\bf FormulaEval} & {\bf ConvFinQA (E)} & {\bf TAT-QA (E)} & {\bf SEC-Num} & {\bf FinCode} & {\bf CodeFinQA} & {\bf CodeTAT-QA} & \\
    & & 0-shot & 0-shot & 3-shot & 3-shot  & 3-shot  & 8-shot  & 3-shot  & 3-shot  &  \\

    \hline
    Falcon           & 7B    & 40.9 & 14.0 & 66.5 & 62.9 & 27.3 & 3.3  & 2.0  & 7.4  & 28.0 \\
    Falcon           & 40B   & 43.9 & 6.0 & 82.8 & 80.6 & 52.3 & 8.3  & 18.4 & 38.5 & 41.3 \\
    \hline
    MPT              & 7B    & 39.7 & 48.0 & 71.7 & 62.1 & 31.7 & 6.6  & 6.6  & 30.4 & 37.1 \\
    MPT              & 30B   & 42.4 & 60.0 & 85.5 & 81.5 & 56.4 & 6.6  & 31.0 & 64.8 & 53.5 \\
    \hline
    StarCoder        & 16B   & 37.9 & 18.0 & 79.7 & 75.0 & 57.8 & 9.9  & 31.2 & 70.2 & 47.5 \\
    \hline
    Llama 2          & 7B    & 41.7 & 52.0 & 86.1 & 83.1 & 56.2 & 7.4  & 21.9 & 37.0 & 48.2 \\
    Llama 2          & 13B   & 42.1 & 52.0 & 88.2 & 88.7 & 61.4 & 6.6  & 33.4 & 65.1 & 54.7 \\
    Llama 2          & 70B   & 44.9 & 80.0 & \underline{92.8} & \textbf{94.4} & 74.7 & 24.0 & 57.3 & 79.1 & 68.4 \\
    \hline
    CodeLlama        & 7B    & 36.5 & 52.0 & 82.4 & 77.4 & 53.1 & 10.7 & 34.0 & 70.9 & 52.1 \\
    CodeLlama        & 13B   & 37.3 & 56.0 & 87.0 & 81.5 & 61.3 & 9.9  & 39.1 & 82.1 & 56.8 \\
    CodeLlama        & 34B   & 40.0 & 70.0 & 88.1 & 83.9 & 67.0 & 17.4 & 52.4 & 81.4 & 62.5 \\
    \hline
    
    Mistral          & 7B    & 44.4 & 66.0 & 91.5 & 87.9 & 65.1 & 15.7 & 48.8 & 75.0 & 61.8 \\
    Mixtral          & 8x7B  & 47.1 & \underline{94.0} & 92.8 & 91.1 & 73.1 & 19.0 & 58.5 & 83.9 &  69.9\\

    \hline
    GPT 3*            & 175B  & 47.9 & 82.0 & 91.6 & \underline{91.9} & \underline{77.4} & 26.5 & 61.5 & 84.7 & 72.3 \\
    GPT 3.5*          & -  & \underline{60.3} & 76.0 & 92.4 & 84.2 & 76.0 & \underline{36.1} & \underline{67.5} & \underline{87.6} & \underline{72.5} \\
    GPT 4*            & -    & \textbf{80.1} & \textbf{100.0} & \textbf{94.0} & 90.3 & \textbf{79.3} & \textbf{63.6} & \textbf{78.8} & \textbf{90.6} & \textbf{84.6} \\
    \hline
\end{tabular}
}
\caption{Performance of state-of-the-art models on BizBench in zero-shot and few-shot in-context learning settings. The best performance for each task is in \textbf{bold.} The second-best performance is \underline{underlined}. Closed-source models are represented by asterisks (*), whereas all other models are open-source.
}\label{tab:fewshot-result}
\end{table*}

There are two main types of functions within this task: standalone functions and class functions. 
The standalone functions represent common financial formulas, such as computing the simple interest rate accrual on a loan. 
Many formulae involve reasoning about the structured relationships between a common set of items, such as computing EBITDA or Net Income from a balance sheet. 
To evaluate these types of formulae, we implement shared classes that represent financial documents (Balance Sheet, Income Statement, Statement of Cash Flows) with attributes representing items that you might find within these documents.

\newtcolorbox{tcolorbox-code}{%
    colback=C_SECOND_SOFT, %
    frame=none,
    colframe=C_SECOND, %
    sharp corners, %
    left=-2pt, right=-2pt, top=0pt, bottom=0pt,
    boxrule=0pt,frame hidden
}

\lstset{numberstyle=\ttfamily}
\begin{figure}
\centering
\begin{tcolorbox}[enhanced jigsaw,colback=white, left=7pt, right=0pt, top=0pt, bottom=0pt, title=FormulaEval]
\vspace{-2mm}
\begin{lstlisting}[backgroundcolor=\color{white},
    rulecolor=\color{black},
    rulesepcolor=\color{gray},
    numbersep=0pt,
    numbers=left,
    ]
def real_rate_of_return(nominal_rate: float, inflation_rate: float) -> float:
    """
    Computes the "real" rate of return, the rate 
    that takes the corresponding rate of 
    inflation into account.
    
    Parameters
    ----------
    nominal_rate: float
    inflation_rate: float
    Returns
    -------
     The "real" rate of return: float
    """
    \end{lstlisting}

    \vspace{-3mm}
    \begin{lstlisting}[backgroundcolor=\color{C_BASE_SOFTER}, 
    rulecolor=\color{black},
    rulesepcolor=\color{gray},
    numbers=none,
    ]
    return ((1 + nominal_rate) / 
                (1 + inflation_rate)) - 1
    \end{lstlisting}
\end{tcolorbox}
\caption{Truncated example from \textbf{FormulaEval}. The model is prompted with the function signature and docstring and generates the code \textcolor{C_BASE}{highlighted in  cyan}.}
\label{fig:formulaeval}
\end{figure}


The model is given a function stub including a docstring and type hints.
For the functions that are part of a class, the model is also given the class definition.
The model is then tasked with completing the implementation of the function. An example of task input and expected output is shown in Figure \ref{fig:formulaeval}. 
In total, we collected 50 functions: 14 of which are standalone functions and 36 functions coming from four class implementations.

Evaluation is done by synthesized unit testing. 
Greedy model-generated and gold function implementations are compared by checking their outputs on 100 randomly sampled inputs.
If their outputs match on all the inputs, we consider the model-generated function implementation to be correct. We report the overall accuracy on this task.

\section{Few-Shot Experiments}


We benchmark state-of-the-art open-source and proprietary models on the BizBench dataset using the EleutherAI LM Evaluation harness \citep{gao_eleutherailm-evaluation-harness_2022}, establishing how performance ranges across model sizes and pretraining strategies. 
For large models, we provide zero- and few-shot results. 
We evaluate {\bf Falcon} \cite{penedo2023refinedweb}, {\bf MPT} \cite{mosaicml2023mpt}, {\bf StarCoder} \cite{li2023starcoder},  {\bf Llama-2} and {\bf CodeLlama} \cite{touvron2023llama}, {\bf Mistral/Mixtral} \cite{jiang2023mistral,jiang2024mixtral}, 
 and OpenAI's {\bf GPT}(s) \cite{brown2020language,openai2023gpt4}. Table \ref{tab:fewshot-result} shows the models and their sizes. 

We were unable to benchmark pre-existing financial LLMs due to three factors: The models we hoped to benchmark were either proprietary (e.g., BloombergGPT \cite{wu2023bloomberggpt}), designed for languages other than English (e.g. BBT-Fin-T5 \citep{fint5} and XuanYuan 2.0 \cite{xuanyuan}) or did not include code in their financial tuning and thus led to very poor results (e.g. FinMA \cite{xie2023pixiu}, FinGPT \cite{yang2023fingpt}, and Fin-LLaMA \cite{Fin-LLAMA}). BizBench requires models to be adept in both finance and coding. We discuss this limitation further in Section~\ref{sec:limits}.

Model size and alignment impact performance: generally, larger models perform better, and models using RLHF and instruction tuning, such as GPT variants, perform best of all. 
There are notable outliers: Mistral 7B outperforms larger models--MPT, Falcon, and Llama 2--across a range of settings.

\subsection{Few-Shot Error Analysis}
We analyze how models err on program synthesis to better understand where they can improve. We categorize model errors into four categories: {\bf extracting} the wrong number from  context, using the wrong {\bf formula}, or code {\bf syntax} errors. Also, some questions are {\bf ambiguous} about the desired answer format (e.g. decimal vs percentage). Our use of strict matching metrics causes some trivial errors.\footnote{Appendix \ref{app:error-analysis} shows examples of these error types.}

\textbf{Upper Limits of Performance.}
GPT-4 is the best performing model on seven of the eight tasks benchmarked here.
For the most difficult task -- FinCode -- GPT-4 achieves 27.5\% absolute improvement over the next best model. 
Despite this, GPT-4 still fails to answer over 36\% of the questions.
We further analyzed these errors to better characterize the current limits in performance.

Of the 44 errors GPT-4 made, there were zero extraction or syntax errors. We deemed 4 errors to be due to potentially ambiguous questions. Of the formulaic errors, one error was the result of answering in the wrong units (trillions of dollars rather than billions of dollars) and two were more basic math mistakes (Error counting number of months, and an error with order of operations). The remaining 37 errors all came from limits in GPT-4's business and financial knowledge. Examples of these errors can be found in Appendix~\ref{app:error-analysis}. This suggests that improving LLMs business and financial background knowledge is key to improving performance.

\textbf{Comparing Code and Language Models.}
Llama-2 and CodeLlama models allow us to measure the benefits of further code training. This additional training results in improvements on our financial program synthesis benchmarks (by an absolute 16.4\% and 8.6\% for 7B and 13B respectively) albeit at the trade-off of slightly worse quantity extraction and domain knowledge performance.

To further understand these differences in performance, we categorize CodeFinQA errors found in 50 incorrect outputs from CodeLlama-34B, as well as 25 from each of CodeLlama-7B and Llama-2-7B.\footnote{Full error counts can be found in Table~\ref{tab:error} in the Appendix} The largest category of errors are numerical extraction errors, followed by incorrect formula errors. We find no clear difference in the error type distribution arising from parameters (CodeLlama-34B vs. CodeLlama-7B) or code pretraining (CodeLlama-7B vs. Llama-2-7B). 
Qualitatively, there are problems with scaling, e.g., confusing millions with billions or ratios with percentages. The models struggle with the semantics of gains and losses, particularly the sign of financial values. For example, the following all indicate a loss: \textit{``loss of \$100''}, \textit{``-\$100''}, \textit{``(\$100)''}.

Models pretrained on language-only versus code solve different sets of questions. 
Llama-2-70B correctly answers 33\% of the questions that CodeLlama-34B fails on; and CodeLlama-34B correctly answers 25\% of those Llama-2-70B fails on. An oracle mixture of these models would achieve an accuracy of 68\% on the CodeFinQA task. 

To analyze the generated code complexity, we measure output length in lines and the number of parentheses for Llama-2-70B on FinCode compared to the values for the gold data shown in Table~\ref{tab:code-complexity}. While the average gold answer contains 7 lines of code, Llama-2-70B outputs contain 6.2, with incorrect responses at 6.4 and correct responses at 5.8 lines. Llama-2-70B may only answer questions where simpler code suffices. However, looking at the number of parentheses we find that Llama-2-70B outputs contain on average 4.0 compared to 4.6 in the gold. Incorrect outputs have an average of 3.9 and correct outputs 4.2. Llama-2-70B may write more structured code when it understands the question better. Further analysis is needed to confirm this hypothesis. 

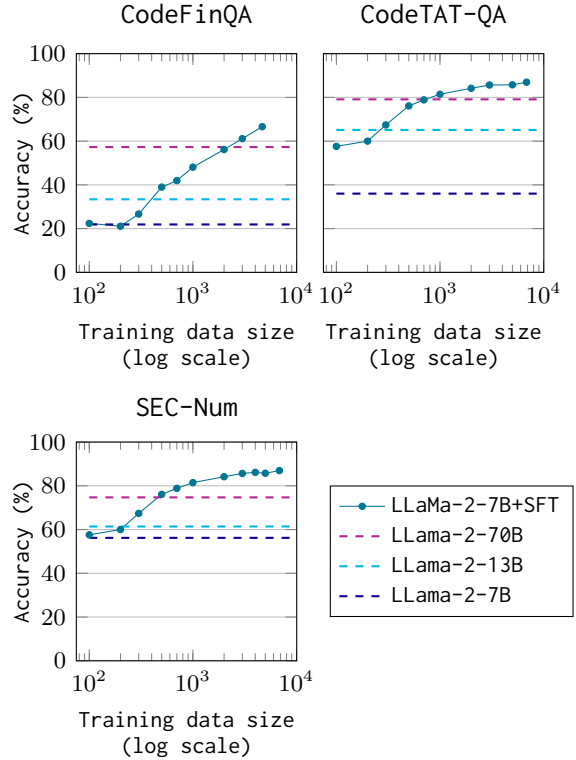
\begin{figure}
    \centering

\ttfamily
\begin{tikzpicture}
    \begin{groupplot}[group style={columns=2},
      view={0}{90},
      width=2.9cm,
      height=2.9cm,
      scale only axis,
      xmode=log,
    log ticks with fixed point,
    log basis x={10},
         group style={
        group size=2 by 2,
        horizontal sep=10pt,
        vertical sep=2.25cm,
        x descriptions at=edge bottom,
        y descriptions at=edge left},
      ymin=0, ymax=100,
    legend,
      legend style={at={(1.15,0.2)},anchor=south west,
      font=\footnotesize},
          legend cell align={left},
      ]    
          \nextgroupplot[
              ylabel=Accuracy (\%),
            ylabel style={at={(-0.15,0.5)}, font=\footnotesize},
              title=CodeFinQA,
          xmin=75,xmax=10050,
          ymajorgrids,
        xlabel style={font=\footnotesize,align=center},
        xlabel={Training data size\\ (log scale)},
          xticklabels={, {$10^2$}, $10^3$, $10^4$ },
          ]
              \addplot+[color=C_BASE, mark = *, mark options={fill=C_BASE, scale=0.6}] coordinates { (100, 22.39)(200, 21.09)(300, 26.66)(500, 38.98)(700, 41.94)(1000, 48.10)(2000, 56.16)(3000, 61.14)(4669, 66.59)};
              \addplot+[mark=none, thick, dashed, C_SECOND, mark options={scale=0.6}] coordinates {(100, 57.3) (10000, 57.3)};
              \addplot+[mark=none, thick, dashed, C_THIRD, mark options={scale=0.6}] coordinates {(100, 33.4) (10000, 33.4)};
              \addplot+[mark=none, thick, dashed, C_FOURTH, mark options={scale=0.6}] coordinates {(100, 21.9) (10000, 21.9)};
          \nextgroupplot[
              title=CodeTAT-QA,
          xmin=75,xmax=10050,
          ymajorgrids,
               xlabel style={font=\footnotesize, align=center},
            xlabel={Training data size \\(log scale)},
          xticklabels={, {$10^2$}, $10^3$, $10^4$ },
          ]
              \addplot+[color=C_BASE, mark = *, mark options={fill=C_BASE, scale=0.6}] coordinates { (100, 57.6) (200, 60.0) (300, 67.4) (500, 76.1) (700, 78.85) (1000, 81.45) (2000, 84.15) (3000, 85.65) (5000, 85.75) (6845, 86.95)};
                \addplot+[mark=none, thick, dashed, C_SECOND, mark options={scale=0.6}] coordinates {(100, 79.1) (10000, 79.1)};
              \addplot+[mark=none, thick, dashed, C_THIRD, mark options={scale=0.6}] coordinates {(100, 65.1) (10000, 65.1)};
              \addplot+[mark=none, thick, dashed, C_FOURTH, mark options={scale=0.6}] coordinates {(100, 36.0) (10000, 36.0)};
          \nextgroupplot[
                  ylabel=Accuracy (\%),
            ylabel style={at={(-0.15,0.5)}, font=\footnotesize},
              title=SEC-Num,
              ymajorgrids,
          xmin=75,xmax=10050,
          xticklabels={, {$10^2$}, $10^3$, $10^4$ },
          xlabel={Training data size \\ (log scale)},
          xlabel style={font=\footnotesize, align=center},]
              \addplot+[color=C_BASE, mark = *, mark options={scale=0.6}] coordinates { (100, 57.6)(200, 60.0)(300,  67.4)(500, 76.1)(700, 78.85)(1000, 81.45)(2000, 84.15)(3000, 85.65)(4000, 86.15)(5000, 85.75) (6845, 86.95)};
                \addplot+[mark=none, thick, dashed, C_SECOND, mark options={scale=0.6}] coordinates {(100, 74.7) (10000,74.7)};
              \addplot+[mark=none, thick, dashed, C_THIRD, mark options={scale=0.6}] coordinates {(100, 61.4) (10000, 61.4)};
              \addplot+[mark=none, thick, dashed, C_FOURTH, mark options={scale=0.6}] coordinates {(100, 56.2) (10000, 56.2)};
              
      \legend{LLaMa-2-7B+SFT, LLama-2-70B, LLama-2-13B, LLama-2-7B}
  
    \end{groupplot}
  \end{tikzpicture}
    \caption{Performance of different finetuned Llama-2-7B models using increasing amounts of training data. Few-shot performance of various Llama 2 models is shown for comparison.}
    \label{fig:finetuning}
\end{figure}

\section{Supervised Finetuning} 
We finetune Llama-2-7B in a number of settings to evaluate how the availability of both in-domain and multi-task training data impacts performance.
Specifically, we investigate how different amounts of training data examples impact performance across CodeFinQA, CodeTAT-QA, and SEC-Num. To test this, we downsample the training data from 100 to 5,000 samples. Figure \ref{fig:finetuning} shows the relationship between accuracy and training data size for finetuned Llama-2-7B models across these sizes compared to the few-shot results of the pretrained models. Next, we evaluate Llama-2-7B with multi-task training across these same datasets, CodeFinQA, CodeTAT-QA, and SEC-Num.


\textbf{Analysis.}
Llama-2-7B requires fewer than 500 samples to outperform the Llama-2-13B on all three datasets. To outperform Llama-2-70B, Llama-2-7B needs to be finetuned on only 3,000 samples for CodeFinQA, 1,000 samples for CodeTAT-QA, and only 500 samples for SEC-Num. Notably, when Llama-2-7B is finetuned with the full training datasets, it demonstrates a substantial improvement -- achieving a 9\%, 7\%, and 13\% higher accuracy on CodeFinQA, CodeTAT-QA, and SEC-Num compared to Llama-2-70B -- while incurring only a fraction of the inference cost. 

We study how knowledge is transferred across these three tasks. Table \ref{tab:multitask} shows a comparison between pretraining, supervised finetuning and multitask learning. 
In zero-shot transfer learning setting (e.g. training on SEC-Num and CodeFinQA and evaluating on CodeTAT-QA), the performance on CodeTAT-QA and SEC-Num improved with additional training data from other tasks, while performance on CodeFinQA slightly decreased. When the training data of the test task is included, data from other tasks has varying impact. Compared to SFT, the performance on CodeFinQA and SEC-Num increased consistently across all training mixes, while the performance on CodeTAT-QA decreased consistently.

\begin{table}
    \centering
    \resizebox{0.6\linewidth}{!}{
    \begin{tabular}{ccc|ccc}
        \toprule
       \multicolumn{3}{l|}{\mtr{2}{\bf Source tasks}} & \multicolumn{3}{c}{\bf Target task} \\
       
       & & & $\rightarrow\bigstar$ & $\rightarrow\blacktriangle$ & \bf $\rightarrow\blacksquare$ \\
        \midrule 
        \multicolumn{3}{l|}{No SFT} & 21.9 & 37.0 & 56.2\\
        \midrule
        \multicolumn{3}{l|}{Single} & 62.4  & \bf 86.5  & 85.7 \\
        \midrule
        & $\blacktriangle$ & $\blacksquare$ & 21.2 & 77.3  & 84.7 \\ 
        $\bigstar$ & & $\blacksquare$ & 65.4 & 49.0 & \bf 87.3 \\ 
        $\bigstar$ & $\blacktriangle$ &  & 65.1 & 81.9 & 69.9 \\ 
        $\bigstar$ & $\blacktriangle$ & $\blacksquare$ & \bf 66.6 & 78.3 & 86.1 \\
        \bottomrule
    \end{tabular}
    }
    \caption{Performance of LLaMa 2 7B model finetuned on different combinations of CodeFinQA($\bigstar$), CodeTAT-QA($\blacktriangle$), and SEC-Num($\blacksquare$) in comparison with no supervised-finetuning (row 1) and supervised finetuning on the same task (row 2).}
    \label{tab:multitask}
\end{table}

\section{Related Work}
BizBench focuses on code generation and financial information extraction, complementing other financial NLP benchmark datasets which tend to focus on non-quantitative tasks  \citep{malo2014good, sinha2021impact}, or quantitative trading tasks \citep{soun2022accurate, han2023select}.

\textbf{Financial NLP.}
Non-quantitative tasks like sentiment detection \cite{malo2014good}, named entity recognition \cite{salinas-alvarado-etal-2015-domain}, classification \cite{sinha2021impact}, question answering \cite{maia2018www}, and boundary detection tasks \cite{au2021finsbd} are included in the financial benchmarks FLARE, introduced within PIXIU \cite{xie2023pixiu}, and FLUE \cite{shah-etal-2022-flue}. \citet{Lu2023BBTFinCC} additionally include entity extraction, event extraction, and natural language to SQL task for Chinese. Recently, stock movement prediction \cite{soun2022accurate} and pair trading \cite{han2023select} were introduced as prediction tasks. BizBench complements these benchmarks by specifically focusing on quantitative reasoning via program synthesis, providing the first tasks of this kind within financial NLP.

\textbf{Code generation.} 
The APPS and HumanEval benchmarks test a model's ability to write code from arbitrary natural language specifications \citep{hendrycksapps2021, chen2021evaluating}. \citet{austin2021program} introduce MBPP, a collection of simple programming questions, as well as a Python version of the MathQA dataset introduced in \citet{amini-etal-2019-mathqa}. We take inspiration from this approach to augmenting existing data when creating CodeTAT-QA and CodeFinQA. Like in our work, \citet{Lai2022DS1000} constructs a dataset of code solutions to data science problems which involve using libraries like numpy and pandas. Our coding tasks also involve reading comprehension, quantity manipulation, and financial knowledge. 

\textbf{Quantity extraction.} 
Previous work explores extracting numbers, metric scales and semantic descriptions in financial 
\cite{loukas-etal-2022-finer} and scientific documents \cite{harper-etal-2021-semeval,elazar-etal-2019-large}. Many financial QA datasets require precise quantity extraction such as HybridQA \cite{chen-etal-2020-hybridqa}, TAT-QA \cite{zhu-etal-2021-tat}, MultiHierTT \cite{zhao-etal-2022-multihiertt}, FinQA \cite{chen-etal-2021-finqa}, and ConvFinQA \cite{chen-etal-2022-convfinqa}. Our BizBench benchmark includes three datasets dedicated to testing numerical extraction capabilities, with our novel SEC-Num dataset encompassing values, scales, currency, dates, times, and periods.

\section{Conclusion}
Numerical reasoning is critically important for business and finance, as well as other domains, because small errors can incur large costs. To address this, we introduce BizBench, a benchmark for business and finance that measures models' abilities across three categories of relevant tasks: domain knowledge, quantity extraction, and program synthesis. The benchmark includes eight tasks, five of which provide novel data or meaningful extensions of existing datasets. We focus on program synthesis, as success on these challenging tasks requires strong capabilities in both domain knowledge and quantity extraction. Additionally, we evaluate many state-of-the-art models, illustrating that models need improvement to meet the demands required for reasoning in real-world, high-stakes domains.

\newpage

\section*{Limitations}
\label{sec:limits}
This work presents new and reformulated data for evaluating the financial reasoning capabilities of LLMs. 
As mentioned in Section \ref{sec:ps}, portions of FinCode, CodeFinQA, and CodeTAT-QA were generated by LLMs, namely WizardCoder \cite{luo2023wizardcoder} and a GPT-3 variant, \verb|text-davinci-003|. We assumed that the code is correct if it produced an answer that is identical or approximately identical to our gold truth answer. This approach could provide false positives (Python code that generated correct answers via an incorrect solution) and false negatives (Python code that had correct solutions by generated output that did not match our criterion).
We have not manually reviewed all the data used to train these models, nor have we manually inspected all the model outputs included in BizBench. Therefore, we cannot make claims about the presence or absence of toxic, biased, or personal information in this data.

For tasks that involve providing context data to models, such data primarily consists of public records filed in accordance with strict government regulations. However, even under these regulations, biases may exist within the data. FinKnow and FinCode questions come from Internet sources and may contain bias. FormulaEval was created with assistance of financial professionals, and despite these professionals' expertise, the data could contain errors.

While BizBench aims to be comprehensive, it does not span the entire domains of business and finance. Moreover, BizBench only contains English text, limiting its applicability to finance problems in other languages. Various financial topics, documents, and problems -- especially those concerning private markets -- are not addressed by this benchmark. By evaluating the model's financial knowledge via it's ability to generate code, we constrain the set of models that can be fairly evaluated with our benchmark. Additionally, many financial professionals do not possess coding skills, implying that this may only be an artificial constraint. 

Annotations included in BizBench were done by individuals with financial knowledge known to the authors. They volunteered to annotate and received no payment for their annotations. All annotators were made aware of the intended use of their data and consented to this use. 

We assume that framing QA as program synthesis allows for easier auditing of model reasoning processes. This assumption is supported by the authors' experiences but may not hold in all settings.

\bibliography{anthology,custom}

\appendix

\clearpage

\section{Data Collection}
\label{app:data-collection}
\subsection{FinCode}
\label{app:fin-code-collection}
This code was annotated by a combination of human and LLMs. For the first phase of data collection, we manually annotated a small set of examples. These examples were concatenated together to create a few-shot prompt that was provided to a GPT-3 variant, \verb|text-davinci-003|, which generated candidate programs for the remaining questions. Programs were manually verified before being added to the dataset. This process was additionally used to identify the most challenging problems for further investigation. These problems were sent to financial professionals to solve by hand. These financial professionals were all full-time members of the finance or accounting staff at a fortune 500 company. After we manually converted these solutions into Python code, we repeated our initial bootstrapping process with the new set of examples. In total, we were able to collect 137 question and program pairs. Of these, 46 were written from scratch by financial professionals, and 91 were generated by an LLM and then verified by financial professionals.

\subsection{CodeFinQA}
\label{app:codefinqa-collection}
To produce code solutions for these questions, we convert the FinQA equations into Python programs using a deterministic process. 
This process defines variables with dummy names (i.e., $a,b,c$) for numbers in the provided equation, and reformulates the operations from the equation into valid Python code over these variables. 
We then use a code-generation model -- WizardCoder 15B \cite{luo2023wizardcoder} -- to rewrite the programs to enhance readability. 
The prompt consists of input and output examples. Each input questions is paired with deterministic code, and the outputs are logical, human-written code with appropriate variable naming conventions.

This prompt is used to seed a bootstrapping process for data annotation. 
First, we find the deterministic programs which are most similar to our seed datapoints and rewrite them using the seeds as prompts. 
We verify that the rewritten program executes to produce the correct answer for its question. 
Subsequently, we find the most similar seed or previously rewritten datapoints for the remaining program until convergence. This
processed produced a dataset of 5,513 question, context, code triples. 

\subsection{CodeTAT-QA}
\label{app:codetatqa-collection}
To create the code for CodeTAT-QA, we first convert the tables from text list format to Pandas DataFrames with named columns and rows. 
During this process, we flatten table hierarchy by attaching sub-table headers to corresponding row labels and normalize number representations, mirroring the action of contemporary document processing models.
Information from the table can be accessed through the dataframe by specifying the row and column labels, as shown in Figure~\ref{fig:codetatqa-example}.

To produce programs for each datapoint, we start with a deterministic process to convert the arithmetic derivation provided in TAT-QA into valid Python code which accesses cells in the table using their row and column names.
If this process identifies the values from the derivation in the table and verifies the provided answer, we pass the deterministically-generated program to a bootstrapping process similar to what was used to create CodeFinQA. 
Some derivations cannot be converted to programs which execute to the gold answer, and many questions from TAT-QA can not be treated as functions over DataFrames alone.
This technique was able to produce code for 3,256 of the 16,000 questions of the TAT-QA dataset. 

\section{Error analysis}

\label{app:error-analysis}

The full counts from our error categorization are shown in Table~\ref{tab:error}. We note that many outputs contain multiple error types. 


\begin{table*}[t]
\centering
\begin{tabular}{lccccc}
\hline
{\bf Model} & {\it Samples} & {\bf Extraction} & {\bf Formula} & {\bf Generation} & {\bf Ambiguous Question }\\
\hline
CodeLlama-34B & 50 & 29 & 23 & 2 & 4 \\ \hline
Llama-2-7B  & 25 & 16 & 13 & 2 & 1 \\ 
CodeLlama-7B & 25 & 17 & 13 & 2 & 2 \\
\hline
\end{tabular}
\caption{Errors by category for different models.}
\label{tab:error}
\end{table*}

We show two common error types i.e. incorrect data extraction and question misinterpretation in Figure \ref{fig:codefinqa-error-1} and Figure \ref{fig:codefinqa-error-2}.

We additionally show two errors of GPT-4 on FinCode in Figure~\ref{fig:gpt4_error_one} and Figure~\ref{fig:gpt4_error_two}.

\lstset{numberstyle=\ttfamily}

\begin{figure}[t]
\small
{\bf Question:} {\it Michael Poole is an equity analyst at Dave Associates, a financial services firm. Poole is estimating the firm's cost of equity using the dividend discount model approach. He has learnt that the Gordon's growth model is particularly useful in deriving the required rate of
 return when this approach is used. The company has paid a dividend of \$2.5 per share in the previous year. The current market price per share is \$25. The company's retention rate and return on equity is 40\% and 10\%, respectively. What is the cost of equity using the dividend discount model? Answer as a percentage to two decimal places. }

\begin{tcolorbox}[enhanced jigsaw, colback=white, left=0pt, right=0pt, top=0pt, bottom=0pt, title=GPT-4 Error Case]
\vspace{-2mm}
\begin{lstlisting}[backgroundcolor=\color{white},
    rulecolor=\color{black},
    rulesepcolor=\color{gray},
    % numbers=none,
    ]
  dividend = 2.5
  price_per_share = 25
  retention_rate = 0.4
  return_on_equity = 0.1
  growth_rate = retention_rate * return_on_equity
  
\end{lstlisting} 
\begin{lstlisting}[backgroundcolor=\color{C_SECOND_SOFTER}, 
    rulecolor=\color{black},
    rulesepcolor=\color{gray},
    numbers=none,
    ]
  cost_of_equity = (dividend / price_per_share) * (1 + growth_rate)
  round(cost_of_equity * 100, 2)
    \end{lstlisting}

\begin{lstlisting}[backgroundcolor=\color{C_BASE_SOFTER}, 
    rulecolor=\color{black},
    rulesepcolor=\color{gray},
    numbers=none,
    ]
  next_dividend_per_share = 
    (dividend * (1 + growth_rate))
  cost_of_equity = 
    (next_dividend_per_share / price_per_share) + growth_rate
  round(100.0 * cost_of_equity, 2)
    \end{lstlisting}
\end{tcolorbox}

\caption{GPT-4 Error Sample from FinCode. The model generated code is shown in {\color{C_SECOND_SOFTER}pink} with the ground truth shown in {\color{C_BASE_SOFTER}cyan}. GPT-4 improperly uses the current dividend value rather than the following one. It additionally incorrectly combines the two ratios with multiplication rather than addition. }
\label{fig:gpt4_error_one}
\end{figure}

\begin{figure}[t]
\small
{\bf Question:} {\it Pine Co. purchased land for \$450000 as a factory site. An existing building on the site was razed before construction began. Additional information is as follows: Cost of razing old building \$60000 Title insurance and legal fees to purchase land \$30000 Architect's fees \$9
5000 New building construction cost \$1850000 What amount should Pine capitalize as the cost of the completed factory building? Answer to the nearest dollar.}

\begin{tcolorbox}[enhanced jigsaw, colback=white, left=0pt, right=0pt, top=0pt, bottom=0pt, title=GPT-4 Error Case]
\vspace{-2mm}
\begin{lstlisting}[backgroundcolor=\color{white},
    rulecolor=\color{black},
    rulesepcolor=\color{gray},
    % numbers=none,
    ]
  land_cost = 450000
  razing_cost = 60000
  title_insurance_and_legal_fees = 30000
  architect_fees = 95000
  construction_cost = 1850000 
\end{lstlisting} 
\begin{lstlisting}[backgroundcolor=\color{C_SECOND_SOFTER}, 
    rulecolor=\color{black},
    rulesepcolor=\color{gray},
    numbers=none,
    ]
  total_cost = razing_cost + architect_fees + construction_cost
  round(total_cost)
    \end{lstlisting}

\begin{lstlisting}[backgroundcolor=\color{C_BASE_SOFTER}, 
    rulecolor=\color{black},
    rulesepcolor=\color{gray},
    numbers=none,
    ]
  total_cost = architect_fees + construction_cost
  round(total_cost)
    \end{lstlisting}
\end{tcolorbox}

\caption{GPT-4 Error Sample from FinCode. The model generated code is shown in {\color{C_SECOND_SOFTER}pink} with the ground truth shown in {\color{C_BASE_SOFTER}cyan}. The generated code incorrectly includes the razing costs into the capitalized cost for the building.}
\label{fig:gpt4_error_two}
\end{figure}

\section{Hyperparameters}

To finetune experiments for CodeFinQA, CodeTAT-QA, and SEC-Num, we use Adam optimizer with an initial learning rate of 2e-6. The training process takes 3 epochs with a batch size of 32. The maximum token length is set to 2048. The model is finetuned on 8 x Nvidia A100-80GB GPUs. For both finetuned and pretrained models we use greedy decoding whenever applicable.

\section{Task Examples}
Examples of input and expected output for each Bizbench task is shown in the following tables and figures:
\begin{itemize}
    \item FinKnow: See Table~\ref{tab:finknow-example}
    \item FormulaEval: See Table~\ref{tab:appendix_example_formula_eval}
    \item ConvFinQA Extract: See Table ~\ref{tab:appendix_example_convfinqa} and \citet{chen-etal-2022-convfinqa}
    \item TAT-QA Extract: See Table \ref{tab:appendix_example_tatqa} and \citet{zhu-etal-2021-tat}
    \item SEC-Num: See Table~\ref{tab:appendix_example_sec_num}
    \item FinCode: See Table~\ref{tab:appendix_example_fincode}
    \item CodeFinQA: See Table~\ref{tab:appendix_example_codefinqa}
    \item CodeTAT-QA: See Table~\ref{tab:appendix_example_codetatqa}
\end{itemize}

\newpage

\begin{figure}[t]
\small
{\bf Question:} {\it What portion of total company used area is company owned? }\\

{\bf Table input:}

{\it | Sq. ft. in thousands |  United States | Other countries | Total | \\
| :--- | :--- |:--- |:--- | \\
| Owned | 4530 | 2417 | 6947 | \\
| Leased | 1037 | 1341 | 2378 | \\
| Total | 5567 | 3758 | 9325 |}
\begin{lstlisting}[backgroundcolor=\color{C_SECOND_SOFTER},numbersep=5pt]
# Incorrect program
owned_area = 4530
total_area = 9325
percent_owned = owned_area / total_area
\end{lstlisting} 
\begin{lstlisting}[backgroundcolor=\color{C_BASE_SOFTER},numbersep=5pt]
# Golden program
company_owned_area = 6947
total_company_used_area = 9325
answer = company_owned_area / total_company_used_area
\end{lstlisting}
\caption{An example of extraction error in {\bf CodeFinQA} task. Model-generated incorrect code is {\color{C_SECOND} highlighted in pink} and the golden code is {\color{C_BASE} highlighted in cyan}. In this incorrect solution, the model extracted the owned area in the US (\textit{owned\_area = 4530}) while it should have extracted the total owned area (\textit{company\_owned\_area=6947}).}
\label{fig:codefinqa-error-1}
\end{figure}

\begin{figure}[t]
\small
{\bf Question:} {\it What are the total proceeds from the issuance of employee options during February 2004, in millions?}\\

{\bf Context input:}

{\it In February 2004 , the company issued to eligible employees 1032717 options with an exercise price of \$11.19 per share, the fair market value of the class a common stock on the date of grant ... }\\
{ \it ... the fair value of ATC Mexico plan options granted during 2002 were \$3611 per share as determined by using the Black-Scholes option pricing model. As described in note 11, all outstanding options were exercised in march 2004. }
\begin{lstlisting}[backgroundcolor=\color{C_SECOND_SOFTER},numbersep=5pt]
# Incorrect program
proceeds_from_issuance_of_employee_options = 3611
answer = proceeds_from_issuance_of_employee_options / 1000000
\end{lstlisting} 
\begin{lstlisting}[backgroundcolor=\color{C_BASE_SOFTER},numbersep=5pt]
# Golden program
options_issued = 1032717
exercise_price = 11.19
proceeds = options_issued * exercise_price
answer = proceeds / 1000000
\end{lstlisting}
\caption{An example where the model misunderstands the question and generates an incorrect solution with incorrect values.  Model-generated incorrect code is {\color{C_SECOND} highlighted in pink} and the golden code is {\color{C_BASE} highlighted in cyan}. The model, however, was able to detect the answer must be in millions. }
\label{fig:codefinqa-error-2}
\end{figure}

\newcommand{\myrotate}[2]{\multirow{#1}{*}{\rotatebox{90}{\bf #2}}}

\begin{table*}
\centering
\begin{tabular}{|c|lp{0.81\linewidth}|}
    \hline
    & &    \\
    \myrotate{9}{CFA Economics} & \mtr{2}{Question} &  {\it As a firm increases the quantity of its product produced, the distance between its ATC and AVC curve: }\\ 
    & &    \\
    & \mtr{3}{Choices} & A. starts increasing. \\
    & & B. starts decreasing. \\ 
    & & C. remains constant. \\ 
    & &    \\
    & {\bf Answer} & B \\ 
    & &    \\
    \hline
    & &    \\
    \myrotate{9}{Microeconomics} & \mtr{1}{Question} &  {\it Which of the following is a characteristic of monopolistic competition?} \\ 
    & &    \\
    & \mtr{4}{Choices} & A. P > MC. \\
    & & B. Efficiency. \\ 
    & & C. Mostly price competition. \\ 
    & & D. P =MR.\\ 
    & &    \\
    & {\bf Answer} & A \\ 
    & &    \\
    \hline
    & &    \\
    \myrotate{10}{Ethics} &\mtr{2}{Question} &  {\it The Theory of \underline{\hspace{1cm}} posits that 3 three levels of moral reasoning exist which an individual can engage in to assess ethical issues, dependant on their cognitive capacity.} \\
    & &    \\
    & \mtr{4}{Choices} & A. Egoism \\
    &  & B. Cognitive moral development \\ 
    &  & C. Power distance \\ 
    &  & D. Uncertainty avoidance\\
    & &    \\
    & {\bf Answer} & B \\ 
    & &    \\
    \hline
    & &    \\
    \myrotate{8}{Finance} & \mtr{1}{Question} &  {\it If a company engages in share repurchases, leverage will increase:} \\ 
    & &    \\
    &  \mtr{3}{Choices} & A. only if the repurchase is financed with debt. \\
    &   & B. only if the repurchase is financed with excess cash. \\ 
    &   & C. whether the repurchase is financed with debt or with excess cash. \\ 
    & &    \\
    &  {\bf Answer} & C \\ 
    & &    \\
    \hline
\end{tabular}

\caption{Examples from {\bf FinKnow} across different topics.}
\label{tab:finknow-example}
\end{table*}

\begin{table*}
\small
\begin{tabular}{|p{\linewidth}|}
\hline
\\
\mtcb{1}{Context:} \\
\begin{lstlisting}[backgroundcolor=\color{white},numbersep=5pt,xleftmargin = 2em]
@dataclass
class IncomeStatement:
    """
    Representation of a Income Statement. All attributes are listed in
    dollars.
    """

    # Sales
    revenue: float
    cost_of_goods_sold: float
    administrative_expenses: float
    depreciation: float
    amortization: float
    # Income from non-core operations
    other_investment_income: float
    # Total tax burden in dollars
    taxes: float
    # Total interest expense in dollars
    interest: float
    shares_outstanding: float
    current_share_price: float

    def gross_profit(self):
       return self.revenue - self.cost_of_goods_sold

    def operating_expenses(self):
        return self.administrative_expenses + self.depreciation + self.amortization

    def operating_income(self):
        return self.gross_profit() - self.operating_expenses()

    def ebit(self):
        return self.operating_income() + self.other_investment_income

    def ebitda(self):
        return self.ebit() + self.amortization + self.depreciation

    def pretax_income(self):
        return self.ebit() - self.interest

    def net_income(self):
        return self.pretax_income() - self.taxes

    def market_capitalization(self):
        return self.shares_outstanding * self.current_share_price

    def earnings_per_share(self):
\end{lstlisting}
\\
\mtcb{1}{Output:} \\

\lstset{firstnumber=48}

\begin{lstlisting}[backgroundcolor=\color{C_BASE_SOFTER},numbersep=5pt,xleftmargin = 2em]
        return self.net_income() / self.shares_outstanding
\end{lstlisting}\\
\hline
\end{tabular}
\caption{Example from {\bf FormulaEval}.
The model is given the class definition, docstrings, and some functions (L1-L47) to generate the missing {\color{C_BASE} return statement} (L48).}
\label{tab:appendix_example_formula_eval}
\end{table*}

\begin{table*}
\small
\begin{tabular}{|p{\linewidth}|}
    \hline
    \\
    \mtcb{1}{\bf Context:}\\
    \\
    {\bf DEVON ENERGY CORPORATION AND SUBSIDIARIES}\
    \\
    {\bf NOTES TO CONSOLIDATED FINANCIAL STATEMENTS – (Continued)}\\
    \\
    {\bf Proved Undeveloped Reserves}\\
    The following table presents the changes in Devon’s total proved undeveloped reserves during 2013 (in MMBoe).\\
    \\
    
    | | | | | |\\
    | :--- | :--- | :--- | :--- | :--- |\\
    | 1 |  | U.S. | Canada | Total |\\
    | 2 | proved undeveloped reserves as of december 31 2012 | 407 | 433 | 840 |\\
    | 3 | extensions and discoveries | 57 | 38 | 95 |\\
    | 4 | revisions due to prices | 1 | -10 ( 10 ) | -9 ( 9 ) |\\
    | 5 | revisions other than price | -91 ( 91 ) | 13 | -78 ( 78 ) |\\
    | 6 | conversion to proved developed reserves | -116 ( 116 ) | -31 ( 31 ) | -147 ( 147 ) |\\
    | 7 | proved undeveloped reserves as of december 31 2013 | 258 | 443 | 701 |\\
    \\
    
    At December 31, 2013, Devon had 701 MMBoe of proved undeveloped reserves. This represents a 17 percent decrease as compared to 2012 and represents 24 percent of total proved reserves. Drilling and development activities increased Devon’s proved undeveloped reserves 95 MMBoe and resulted in the conversion of 147 MMBoe, or 18 percent, of the 2012 proved undeveloped reserves to proved developed reserves. Costs incurred related to the development and conversion of Devon’s proved undeveloped reserves were \$1.9 billion for 2013. Additionally, revisions other than price decreased Devon’s proved undeveloped reserves 78 MMBoe primarily due to evaluations of certain U.S. onshore dry-gas areas, which Devon does not expect to develop in the next five years. The largest revisions relate to the dry-gas areas in the Cana-Woodford Shale in western Oklahoma, Carthage in east Texas and the Barnett Shale in north Texas.\\
    \\
    A significant amount of Devon’s proved undeveloped reserves at the end of 2013 related to its Jackfish operations. At December 31, 2013 and 2012, Devon’s Jackfish proved undeveloped reserves were 441 MMBoe and 429 MMBoe, respectively. Development schedules for the Jackfish reserves are primarily controlled by the need to keep the processing plants at their 35,000 barrel daily facility capacity. Processing plant capacity is controlled by factors such as total steam processing capacity, steam-oil ratios and air quality discharge permits. As a result, these reserves are classified as proved undeveloped for more than five years. Currently, the development schedule for these reserves extends though the year 2031.\\
    \\
    {\bf Price Revisions}\\
    \\
    2013 – Reserves increased 94 MMBoe primarily due to higher gas prices. Of this increase, 43 MMBoe related to the Barnett Shale and 19 MMBoe related to the Rocky Mountain area.\\
    \\
    2012 – Reserves decreased 171 MMBoe primarily due to lower gas prices. Of this decrease, 100 MMBoe related to the Barnett Shale and 25 MMBoe related to the Rocky Mountain area.\\
    \\
    2011 – Reserves decreased 21 MMBoe due to lower gas prices and higher oil prices. The higher oil prices increased Devon’s Canadian royalty burden, which reduced Devon’s oil reserves.\\
    \\
    {\bf Revisions Other Than Price}\\
    \\
    Total revisions other than price for 2013, 2012 and 2011 primarily related to Devon’s evaluation of certain dry gas regions, with the largest revisions being made in the Cana-Woodford Shale, Barnett Shale and Carthage area.\\
    \\
    \mtcb{1}{\bf Question: }\\
    \\
    \multicolumn{1}{|c|}{\it What is the balance of proved undeveloped reserves in 2012 in US?}\\ 
    \\
    \hline
    \\
    {\bf Answer: } 407\\
    \\
\hline
\end{tabular}
\caption{Example from  {\bf ConvFinQA}. The titles were bolded for readability of the text. We do not use any additional tokens to mark the title during prompting the models.}
\label{tab:appendix_example_convfinqa}
\end{table*}

\begin{table*}
\footnotesize
\begin{tabular}{|p{\linewidth}|}
    \hline
    \\
    \mtcb{1}{\bf Table:}\\

    | - | 2019\% |  2018\% | 2017\% |\\
    | Rate of inflation | 2.9 | 2.9 | 3.0 | \\
    | Rate of increase in salaries | 2.7  | 2.7 | 2.6 | \\
    | Discount rate | 2.3 | 2.5 | 2.6 |\\
\\
\hline
\\
\mtcb{1}{\bf Question: }\\
\\
\multicolumn{1}{|c|}{\it How much is the 2019 rate of inflation?}\\ 
\\
\hline
\\
{\bf Answer: } 2.9\\
\\
\hline
\\

\mtcb{1}{\bf Question: }\\
\\
\multicolumn{1}{|c|}{\it How much is the 2018 rate of inflation?}\\ 
\\
\hline
\\
{\bf Answer: } 2.9\\
\\
\hline
\\

\mtcb{1}{\bf Question: }\\
\\
\multicolumn{1}{|c|}{\it What is the 2019 average rate of inflation?}\\ 
\\
\hline
\\
{\bf Answer: } 2.9\\
\\
\hline
\\

\mtcb{1}{\bf Question: }\\
\\
\multicolumn{1}{|c|}{\it What is the 2019 average rate of increase in salaries?}\\ 
\\
\hline
\\
{\bf Answer: } 2.7\\
\\
\hline
\\

\mtcb{1}{\bf Question: }\\
\\
\multicolumn{1}{|c|}{\it What is the difference between 2019 average rate of inflation and 2019 average rate of increase in salaries}\\ 
\\
\hline
\\
{\bf Answer: } 0.2\\
\\
\hline

\end{tabular}

\caption{Example from {\bf TAT-QA Extract}. Each table can have multiple associated question-answer pairs.}
\label{tab:appendix_example_tatqa}
\end{table*}

\begin{table*}
\footnotesize
\begin{tabular}{|p{\linewidth}|}
    \hline
    \\
    \mtcb{1}{\bf Context:}\\

    {\bf Contractual Obligations}\\
    The following table summarizes scheduled maturities of the Company’s contractual obligations for which cash flows are fixed and determinable as of June 30, 2022:
    \\\\
    |  |  |  |  |  |  |  |  |  \\
    | --- | --- | --- | --- | --- | --- | --- | --- | \\
    |  |  |  |  |  |  |  |  | \\
    |   |  |   |  | Payments Due in Fiscal |  |   |\\
    | (In millions) |  Total | 2023 |2024 | 2025 | 2026 | 2027 |   Thereafter |\\
    | Debt service (1) |   \$8,151  | \$429  | \$170  |  \$665  |   \$161  |   \$661  | \$6,065  | \\
    | Unconditional purchase obligations (2) |   4,742  |2,852  | 705  | 637  | 132  | 133  | 283  | \\
    | Gross unrecognized tax benefits and interest – current (3) |   2  |  2  |  — |  — |   — |   — |   — |\\
    | Transition Tax payable(4) |  215  |  27  | 42  | 65  | 81  |  —  |  —   |\\
    | Total contractual obligations(5) |  \$13,110  | \$3,310  | \$917  |\$1,367  | \$374  | \$794  | \$6,348  | \\
    |  |  |  |  |  |  |  |  |  \\
    \\
    (1) Includes long-term and current debt and the related projected interest costs. Refer to Note 7 – Leases for information regarding future minimum lease payments relating to the Company’s finance leases. {\bf Interest costs on long-term and current debt} in fiscal 2023, 2024, 2025, {\bf 2026}, 2027 and thereafter are projected to be \$174 million, \$170 million, \$165 million, \$161 million, {\bf \$161 million} and \$1,765 million, respectively. Projected interest costs on variable rate instruments were calculated using market rates at June 30, 2022. \\
    \\

    (2) Unconditional purchase obligations primarily include: royalty payments pursuant to license agreements, inventory commitments, information technology contract commitments, capital expenditure commitments, advertising commitments and third-party distribution commitments. Future royalty and advertising commitments were estimated based on planned future sales for the term that was in effect at June 30, 2022, without consideration for potential renewal periods.\\
    \\
    (3) Refer to Note 9 – Income Taxes for information regarding unrecognized tax benefits. As of June 30, 2022, the noncurrent portion of the Company’s unrecognized tax benefits, including related accrued interest and penalties, was \$73 million. At this time, the settlement period for the noncurrent portion of the unrecognized tax benefits, including related accrued interest and penalties, cannot be determined and therefore was not included.\\ \\
    (4) The Transition Tax may be paid over an eight-year period and this amount represents the remaining liability as of June 30, 2022.\\ \\
    (5) Refer to Note 7 – Leases for information regarding future minimum lease payments relating to the Company’s operating leases.  \\
\\

\mtcb{1}{\bf Label: }\\
\\
\multicolumn{1}{|c|}{\it Projected interest costs on long-term and current debt Due in fiscal 2026}\\ 
\\

\hline
\\
{\bf Answer: } 161\\
    \\
\hline
\end{tabular}

\caption{Example from {\bf SEC-Num}. The titles were bolded to enhance the readability of the text. We do not use any additional tokens to mark the title during prompting the models.}
\label{tab:appendix_example_sec_num}
\end{table*}

\begin{table*}
\footnotesize
\begin{tabular}{|p{\linewidth}|}
    \hline
\\
\mtcb{1}{\bf Question: }\\

\\{\it Mill Co. reported pretax income of \$152500 for the year ended December 31. During the year-end audit the external auditors discovered the following errors: Ending inventory \$30000 Overstated Depreciation expense \$64000 What amount should Mill report as the correct pretax income for the year ended December 31? Answer to the closest dollar.}\\ 
\\

\mtcb{1}{\bf Program:}\\
\begin{lstlisting}[backgroundcolor=\color{C_BASE_SOFTER},numbersep=5pt,xleftmargin = 2em]
reported_income = 152500 
ending_inventory = 30000 
depreciation_expense = 64000 
pretax_income = reported_income - ending_inventory - depreciation_expense 
round(pretax_income)
\end{lstlisting}
\\
\hline
\\
{\bf Answer: } 58500\\
    \\
\hline
\end{tabular}

\vspace{2em}

\begin{tabular}{|p{\linewidth}|}
    \hline
\\
\mtcb{1}{\bf Question: }\\

\\{\it Suppose that the market price of Company X is \$45 per share and that of Company Y is \$30. If X offers three-fourths a share of common stock for each share of Y, the ratio of exchange of market prices would be: Answer to three decimal places.}\\ 
\\

\mtcb{1}{\bf Program:}\\
\begin{lstlisting}[backgroundcolor=\color{C_BASE_SOFTER},numbersep=5pt,xleftmargin = 2em]
price_x = 45 
price_y = 30 
exchange_ratio = 3.0 / 4.0 
ratio_of_exchange_market_prices = price_x / price_y * exchange_ratio 
round(ratio_of_exchange_market_prices, 3)
\end{lstlisting}
\\
\hline
\\
{\bf Answer: } 1.125\\
    \\
\hline
\end{tabular}

\vspace{2em}

\begin{tabular}{|p{\linewidth}|}
    \hline
\\
\mtcb{1}{\bf Question: }\\

\\{\it A bond is currently priced at 89.187 per 100 par value. If yields increase by 10bp, the value of bond falls to 88.215. However, if yields decrease by the same amount the value of the bond rises to 90.237. What will the approximate modified duration for the bond be? Answer to two decimal places.}\\ 
\\

\mtcb{1}{\bf Program:}\\
\begin{lstlisting}[backgroundcolor=\color{C_BASE_SOFTER},numbersep=5pt,xleftmargin = 2em]
price_1 = 89.187 
price_2_rise = 90.237 
price_2_fall = 88.215 
yield_rise = 0.001 
yield_fall = -0.001 
modified_duration = (price_2_rise - price_2_fall) / (2 * price_1 * yield_rise) 
round(modified_duration, 2)
\end{lstlisting}
\\
\hline
\\
{\bf Answer: } 11.34\\
    \\
\hline
\end{tabular}
\caption{Examples from {\bf FinCode}. The model needs to generate the code {\color{C_BASE} highlighted in cyan.}}
\label{tab:appendix_example_fincode}
\end{table*}

\begin{table*}
\small
\begin{tabular}{|p{\linewidth}|}
    \hline
    \\
    \mtcb{1}{\bf Context:}\\
    \\
    {\bf NOTES TO CONSOLIDATED FINANCIAL STATEMENTS} (continued)\\
    {\bf ACE Limited and Subsidiaries} \\
    \\
    The following table shows changes in the Company’s restricted stock for the years ended December 31, 2007, 2006, and 2005:
    \\
    \\
    {\bf TABLE} \\
    | | Number of Restricted Stock |  Weighted Average Grant Date Fair Value | \\
    | Unvested restricted stock December 31 2005 | 3488668 |   \$41.26 |\\
    | Granted     | 1632504 | \$56.05 | \\
    | Vested and issued | (1181249) |   \$40.20 | \\
    | Forfeited | (360734) | \$44.04 | \\
    | Unvested restricted stock December 31 2006 | 3579189 | \$48.07 | \\
    | Granted | 1818716 | \$56.45 | \\
    | Vested and issued | (1345412) | \$44.48 | \\
    | Forfeited | (230786) | \$51.57 | \\
    | Unvested restricted stock December 31 2007 | 3821707 | \$53.12 | \\
    | Granted |  1836532 | \$59.84 | \\
    | Vested and issued | (1403826) | \$50.96 | \\
    | Forfeited | (371183) | \$53.75 | \\
    | Unvested restricted stock December 31 2008 | 3883230 | \$57.01 | \\
    {\bf END TABLE} \\
    \\
    Under the provisions of FAS 123R, the recognition of deferred compensation, a contra-equity account representing the amount of unrecognized restricted stock expense that is reduced as expense is recognized, at the date restricted stock is granted is no longer permitted. Therefore, upon adoption of FAS 123R, the amount of deferred compensation that had been reflected in unearned stock grant compensation was reclassified to additional paid-in capital in the company 2019s consolidated balance sheet. \\
    \\
    {\bf  Restricted stock units }\\
    The Company’s 2004 LTIP also provides for grants of other awards, including restricted stock units. The Company generally grants restricted stock units with a 4-year vesting period, based on a graded vesting schedule. Each restricted stock unit represents the Company’s obligation to deliver to the holder one share of Ordinary Shares upon vesting. During 2007, the Company awarded 108,870 restricted stock units to officers of the Company and its subsidiaries with a weighted-average grant date fair value of \$56.29. During 2006, 83,370 restricted stock units, with a weighted-average grant date fair value of \$56.36, were awarded to officers of the Company and its subsidiaries. During 2005, 80,550 restricted stock units, with a weighted-average grant date fair value of \$44.59, were awarded to officers of the Company and its subsidiaries.\\
    \\
    The Company also grants restricted stock units with a 1-year vesting period to non-management directors. Delivery of
    Ordinary Shares on account of these restricted stock units to non-management directors is deferred until six months after the date
    of the non-management directors’ termination from the Board. During 2007, 29,676 restricted stock units were awarded to non-management directors. These units will vest in May 2008. During 2006, 23,092 restricted stock units were awarded to non-management directors. These units vested in May 2007. During 2005, 26,186 restricted stock units were awarded to non-management directors. These units vested in May 2006.\\
    \\
    {\bf ESPP }\\
    The ESPP gives participating employees the right to purchase Ordinary Shares through payroll deductions during consecutive
    “Subscription Periods.” Annual purchases by participants are limited to the number of whole shares that can be purchased by an amount equal to ten percent of the participant’s compensation or \$25,000, whichever is less. The ESPP has two six-month Subscription Periods, the first of which runs between January 1 and June 30 and the second of which runs between July 1 and December 31 of each year. \\
    \mtcb{1}{\bf Question: }\\
    \\
    \multicolumn{1}{|c|}{\it What is the net change in the number of unvested restricted stocks in 2007?}\\ 
    \\
    \mtcb{1}{\bf Program:}\\
    \begin{lstlisting}[backgroundcolor=\color{C_BASE_SOFTER},numbersep=5pt,xleftmargin = 2em]
    unvested_restricted_stocks_2006 = 3579189 
    unvested_restricted_stocks_2007 = 3821707 
    unvested_restricted_stocks_2008 = 3883230 
    change_in_unvested_restricted_stocks = unvested_restricted_stocks_2007 - unvested_restricted_stocks_2006 
    answer = change_in_unvested_restricted_stocks
    \end{lstlisting}
    \\
    \hline
    \\
    {\bf Answer: } 242518\\
        \\
    \hline
\end{tabular}
\caption{Examples from {\bf CodeFinQA}. The titles were made bold to enhance the readability of the text. We do not use any additional tokens to mark the title during prompting the models. The model only needs to generate the code {\color{C_BASE} highlighted in cyan.}}
\label{tab:appendix_example_codefinqa}
\end{table*}

\begin{table*}
\footnotesize
\begin{tabular}{|p{\linewidth}|}
    \hline
    \\
    \mtcb{1}{\bf Table:}\\

    | (in millions) | March 29, 2019 |  March 30, 2018 |\\
    | Net investment hedges -- Foreign exchange forward contracts sold | 116 | - | \\
    | Balance sheet contracts -- Foreign exchange forward contracts purchased | 963 | 697 | \\
    | Balance sheet contracts -- Foreign exchange forward contracts sold | 122  | 151 | \\
\\
\mtcb{1}{\bf Question: }\\
\\
\multicolumn{1}{|c|}{\it What is the change between Foreign exchange forward contracts purchased for March 29, 2019 and March 30, 2018?}\\ 
\\
\mtcb{1}{\bf Program:}\\
\begin{lstlisting}[backgroundcolor=\color{white},numbersep=5pt,xleftmargin = 2em]
table = {
    "Net investment hedges -- Foreign exchange forward contracts sold": {
        "March 29, 2019": 116,
        "March 30, 2018": "-",
    },
    "Balance sheet contracts -- Foreign exchange forward contracts purchased": {
        "March 29, 2019": 963,
        "March 30, 2018": 697,
    },
    "Balance sheet contracts -- Foreign exchange forward contracts sold": {
        "March 29, 2019": 122,
        "March 30, 2018": 151,
    },
}
df = pandas.DataFrame(data=table)
\end{lstlisting}

\lstset{firstnumber=16}

\begin{lstlisting}[backgroundcolor=\color{C_BASE_SOFTER},numbersep=5pt,xleftmargin=2em]
march_29_2019_foreign_exchange_balance_sheet_contract = \
    df["Balance sheet contracts -- Foreign exchange forward contracts purchased"]["March 29, 2019"]
march_30_2018_foreign_exchange_balance_sheet_contract = \
    df["Balance sheet contracts -- Foreign exchange forward contracts purchased"]["March 30, 2018"]
answer = march_29_2019_foreign_exchange_balance_sheet_contract - \
    march_30_2018_foreign_exchange_balance_sheet_contract
\end{lstlisting}
\\
\hline
\\
{\bf Answer: } 266\\
    \\
\hline
\end{tabular}

\vspace{1em}

\begin{tabular}{|p{\linewidth}|}
    \hline
    \\
    \mtcb{1}{\bf Table:}\\

    |          | 2019 | 2018 | 2017 |\\
    | Domestic | 204.2 | 140.3 | 56.0 |\\
    | Foreign | 11.8 | 19.9 | 14.2 | \\
    | Income before income taxes | 216.0 | 160.2 | 70.2 |\\
\mtcb{1}{\bf Question: }\\
\\
\multicolumn{1}{|c|}{\it What was the change in Foreign in 2019 from 2018?}\\ 
\\
\mtcb{1}{\bf Program:}\\
\begin{lstlisting}[backgroundcolor=\color{white},numbersep=5pt,xleftmargin = 2em]
table = {
    "Domestic": {
        "2019": 204.2,
        "2018": 140.3,
        "2017": 56.0
    },
    "Foreign": {
        "2019": 11.8,
        "2018": 19.9,
        "2017": 14.2
    },
    "Income before income taxes": {
        "2019": 216.0,
        "2018": 160.2,
        "2017": 70.2
    }
}
df = pandas.DataFrame(data=table)
\end{lstlisting}

\lstset{firstnumber=19}

\begin{lstlisting}[backgroundcolor=\color{C_BASE_SOFTER},numbersep=5pt,xleftmargin = 2em]
foreign_2019 = df["Foreign"]["2019"]
foreign_2018 = df["Foreign"]["2018"]
answer = foreign_2019 - foreign_2018
\end{lstlisting}
\\
\hline
\\
{\bf Answer: } -8.1\\
    \\
\hline
\end{tabular}

\caption{Examples from {\bf CodeTAT-QA} dataset. The model only needs to generate the code {\color{C_BASE} highlighted in cyan.}}
\label{tab:appendix_example_codetatqa}
\end{table*}

\begin{table*}
\small
\begin{tabular}{|p{\linewidth}|}
    \hline
    \\
    
    \mtcb{1}{\bf Text}\\
    \\
    
    Contractual Obligations\\
    The following table summarizes scheduled maturities of the Company’s contractual obligations for which cash flows are fixed and determinable as of June 30, 2022:
    \\
    |  |  |  |  |  |  |  |  |  \\
    | --- | --- | --- | --- | --- | --- | --- | --- | \\
    |  |  |  |  |  |  |  |  | \\
    |   |  |   |  | Payments Due in Fiscal |  |   |\\
    | (In millions) |  Total | 2023 |2024 | 2025 | 2026 | 2027 |   Thereafter |\\
    | Debt service (1) |   \$8,151  | \$429  | \$170  |  \$665  |   \$161  |   \$661  | \$6,065  | \\
    | Unconditional purchase obligations (2) |   4,742  |2,852  | 705  | 637  | 132  | 133  | 283  | \\
    | Gross unrecognized tax benefits and interest – current (3) |   2  |  2  |  — |  — |   — |   — |   — |\\
    | Transition Tax payable(4) |  215  |  27  | 42  | 65  | 81  |  —  |  —   |\\
    | Total contractual obligations(5) |  \$13,110  | \$3,310  | \$917  |\$1,367  | \$374  | \$794  | \$6,348  | \\
    |  |  |  |  |  |  |  |  |  \\
    \\
    (1) Includes long-term and current debt and the related projected interest costs. Refer to Note 7 – Leases for information regarding future minimum lease payments relating to the Company’s finance leases. Interest costs on long-term and current debt in fiscal 2023, 2024, 2025, 2026, 2027 and thereafter are projected to be \$174 million, \$170 million, \$165 million, \$161 million, \$161 million and \$1,765 million, respectively. Projected interest costs on variable rate instruments were calculated using market rates at June 30, 2022. \\
    \\

    (2) Unconditional purchase obligations primarily include: royalty payments pursuant to license agreements, inventory commitments, information technology contract commitments, capital expenditure commitments, advertising commitments and third-party distribution commitments. Future royalty and advertising commitments were estimated based on planned future sales for the term that was in effect at June 30, 2022, without consideration for potential renewal periods.\\
    \\
    (3) Refer to Note 9 – Income Taxes for information regarding unrecognized tax benefits. As of June 30, 2022, the noncurrent portion of the Company’s unrecognized tax benefits, including related accrued interest and penalties, was \$73 million. At this time, the settlement period for the noncurrent portion of the unrecognized tax benefits, including related accrued interest and penalties, cannot be determined and therefore was not included.\\
    ... \\
    \mtcb{1}{\bf Table Preview}\\
    \begin{minipage}{\linewidth}
        \includegraphics[width=\linewidth]{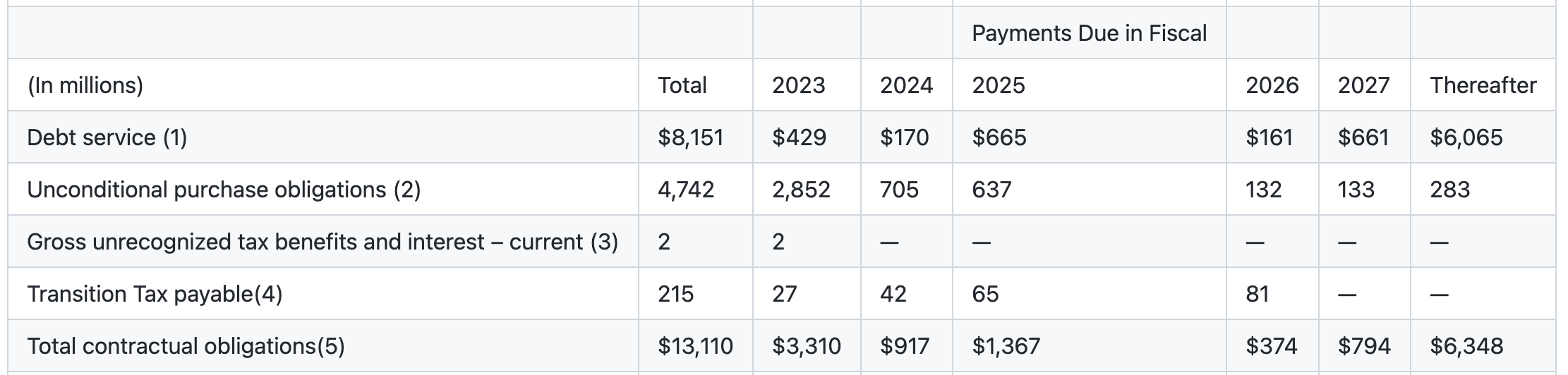}
    \end{minipage}
    \\
    \\
    \mtcb{1}{\bf SEC-annotated value-label pairs}\\
    \{"value": "161", "label": "Projected interest costs on long-term and current debt Due in fiscal 2026"\} \\
    \{"value": "65", "label": "Transition Tax payable, Payments Due in Fiscal 2025"\} \\
    \{"value": "283", "label": "Unconditional purchase obligations, Payments Due in Fiscal Thereafter"\} \\
    \{"value": "170", "label": "Projected interest costs on long-term and current debt Due in fiscal 2024"\} \\
    \{"value": "215", "label": "Transition Tax payable"\} \\
    \{"value": "637", "label": "Unconditional purchase obligations, Payments Due in Fiscal 2025"\} \\
    \{"value": "42", "label": "Transition Tax payable, Payments Due in Fiscal 2024"\} \\
    \{"value": "429", "label": "Debt service, Payments Due in Fiscal 2023"\} \\
    \{"value": "81", "label": "Transition Tax payable, Payments Due in Fiscal 2026"\} \\
    \{"value": "4,742", "label": "Unconditional purchase obligations"\} \\
    \{"value": "132", "label": "Unconditional purchase obligations, Payments Due in Fiscal 2026"\} \\
    \{"value": "27", "label": "Transition Tax payable, Payments Due in Fiscal 2023"\} \\
    \{"value": "665", "label": "Debt service, Payments Due in Fiscal 2025"\} \\
    \{"value": "161", "label": "Debt service, Payments Due in Fiscal 2026"\} \\
    \{"value": "133", "label": "Unconditional purchase obligations, Payments Due in Fiscal 2027"\} \\
    \hline  
\end{tabular}
\caption{Original annotation from SEC filings for the ticker symbol {\bf EL} (2022/06/30). Pairs with identical label are discarded. We present a preview of the table in markdown format for better readability.}
\label{tab:appendix-sec-filing}
\end{table*}

\clearpage

\end{document}